\documentclass{article}
\usepackage{spconf,amsmath,graphicx}

\usepackage{epsfig}
\usepackage{graphicx}
\usepackage{amsmath}
\usepackage{amssymb}

\usepackage{stackengine}
\usepackage{xcolor}
\usepackage{amsfonts}
\usepackage{caption}
\usepackage{algorithm}
\usepackage{algorithmic}
\usepackage{array} 
\usepackage{multirow}
\usepackage{color}    
\usepackage{balance}
\usepackage{cite}
\usepackage{balance}

\definecolor{ultramarine}{RGB}{18, 10, 143} 
\definecolor{xgreen}{RGB}{0,192,0} 
\definecolor{bblue}{RGB}{0,0,128} 
\usepackage[pagebackref=true,breaklinks=true,letterpaper=true,colorlinks,bookmarks=false,citecolor=bblue]{hyperref}

\newcommand{\capname}{Style Projection}
\newcommand{\methodname}{Style Projection}
\newcommand{\abbr}{Style Projection}

\title{Parameter-Free Style Projection for Arbitrary Image Style Transfer}

\name{\begin{tabular}{c}Siyu Huang$^1$\sthanks{Work done when Siyu Huang and Jun Huan were at Baidu Research.},
        Haoyi Xiong$^2$,
        Tianyang Wang$^3$, 
        Bihan Wen$^4$, 
        \\
        \textit{Qingzhong Wang}$^2$, 
        \textit{Zeyu Chen}$^2$, 
        \textit{Jun Huan}$^{5*}$,
        \textit{Dejing Dou}$^2$
        \end{tabular}
        }
\address{$^1$Harvard University, USA ~~~~~~~
$^2$Baidu Research, China  ~~~~~~~
$^3$Austin Peay State University, USA \\
$^4$Nanyang Technological University, Singapore ~~~~~~~
$^5$Amazon, USA
}

\begin{document}
\maketitle
\begin{abstract}
Arbitrary image style transfer is a challenging task which aims to stylize a content image conditioned on arbitrary style images. In this task the feature-level content-style transformation plays a vital role for proper fusion of features. Existing feature transformation algorithms often suffer from loss of content or style details, non-natural stroke patterns, and unstable training. To mitigate these issues, this paper proposes a new feature-level style transformation technique, named Style Projection, for parameter-free, fast, and effective content-style transformation. This paper further presents a real-time feed-forward model to leverage Style Projection for arbitrary image style transfer, which includes a regularization term for matching the semantics between input contents and stylized outputs. Extensive qualitative analysis, quantitative evaluation, and user study have demonstrated the effectiveness and efficiency of the proposed methods.
\end{abstract}


\vspace{-0.05em}
\section{Introduction}

Arbitrary image style transfer \cite{gatys2016image} is a very challenging task which aims to synthesize artistic images conditioned on arbitrary styles, where the content-style feature transformation plays a vital role and a series of content-style transformation algorithms have been proposed in the literature \cite{heeger1995pyramid, 9117149, de1997multiresolution, portilla2000parametric, cheng2017video}. The widely-used normalization-based methods, including instance normalization (IN) \cite{in}, conditional IN (CIN) \cite{cin}, and adaptive IN (AdaIN) \cite{adain,spade,stylegan}, generally normalize the style of content features to style features. However, only limited style information, \textit{i.e.}, the first-order statistics of style features, is injected into the content features. Another widely-adopted feature-level style transformation method is whitening and coloring transformation (WCT) \cite{wct} which completely peels off the style information of content features then recovers with the target style information. Although the whitening operation was designed to remove style information, it generates artifacts that content details are also removed unintentionally, leading to unpleasant synthesized results. Both normalization-based methods and WCT-based methods cannot well balance the content preserving and style transferring.

In this paper, we propose a new feature-level style transformation technique, named~\methodname, which is inspired by the order statistics \cite{david2004order,pitas1992order}. The order statistics-based image processing filters, such as max/min/median filters and ranked-order filters, can preserve fine content details of images in applications of noise detection \cite{noise_detection1,noise_detection2}, image denoising \cite{image_denoising}, and object detection \cite{object_detection}. In this work, we exploit order statistics for deep neural network-based content-style feature transformation for the first time, by considering content/style information as order statistics/scalar values of features, respectively. In a very simple manner, \methodname~reorders the style features according to the order of the content features, such that the correlation of feature values (shapes, textures, etc.) is provided by the content features, while the color information is provided by the style features, enabling a reasonable content-style fusion.

We further present a learning-based feed-forward model to leverage \methodname~for arbitrary image style transfer in real time. The model works in an encoder-decoder fashion that stylizes a content image based on arbitrarily given style images. In experiments, we conduct extensive studies including quantitative evaluation, qualitative analysis, ablation studies, and user study, to comprehensively validate the efficacy of our style transfer framework. The contributions of this paper are summarized as follows: (1) We present a new parameter-free feature-level transformation technique, named~\methodname, for fast yet effective content-style feature fusion; (2) We present a real-time feed-forward method for arbitrary style transfer, including a KL divergence loss for further matching the semantics between input contents and stylized output; (3) We demonstrate the effectiveness and efficiency of our proposed methods through extensive empirical studies.

\section{Method}

\subsection{Arbitrary Image Style Transfer}
The goal of arbitrary image style transfer is to stylize a content image conditioned on an arbitrary style image. To address this task, we introduce a learning-based feed-forward style transfer model. Following the image style transfer practice in previous literature \cite{adain,wct}, we embed content and style images into content and style features with an image encoder $E$. As discussed above, the critical component of our style transfer model is \methodname~algorithm, which is a parameter-free feature transformation approach for proper fusion of content and style features. Based on \methodname, the content and style features are fused and then reconstructed as a new stylized image with an image decoder $D$. Despite the simplicity of the model, it is effective and there is a leeway to incorporate useful tricks, for instance the semantic matching regularization of KL divergence, to further boost the style transfer performance. In the following sections, we first introduce \methodname~algorithm and then discuss the learning objectives of our style transfer model.

\subsection{\capname~Algorithm}

\begin{algorithm}[t] 
\caption{Parameter-Free \abbr.}
\label{alg}
\begin{algorithmic}[1]
\REQUIRE 
content feature map $x \in \mathbb{R}^{C \cdot H \cdot W}$ \\
style feature map $y \in \mathbb{R}^{C \cdot H \cdot W}$
\ENSURE stylized feature map $z \in \mathbb{R}^{C \cdot H \cdot W}$
\STATE $\vec x \in \mathbb{R}^{C \cdot V} \gets $ reshape $x$; \\
$\vec y \in \mathbb{R}^{C \cdot V} \gets $ reshape $y$;
\STATE index $d_x$, values $\vec x_r \gets$ sort $\vec x$ along $V$; \\
index $d_y$, values $\vec y_r \gets$ sort $\vec y$ along $V$;
\STATE $\vec z \left[:,i\right] \gets \vec y_r \left[:,d_x\left[i\right]\right]$, ~~$i=1, 2, \dots, V$;
\STATE $z \in \mathbb{R}^{C \cdot H \cdot W} \gets$ reshape $\vec z$.
\end{algorithmic}
\end{algorithm}


In this work, we introduce \methodname~which is a simple, fast, yet effective algorithm for content-style fusion. Given a content feature map $x \in \mathbb{R}^{C \cdot H \cdot W}$ and a style feature map $y \in \mathbb{R}^{C \cdot H \cdot W}$, where $C$, $H$, $W$ are channel, height, and width respectively, we firstly vectorize $x$ and $y$ across dimension $H$ and $W$ to obtain features $\vec x \in \mathbb{R}^{C \cdot V}$ and $\vec y \in \mathbb{R}^{C \cdot V}$, where $V = H \cdot W$. Then we compute rankings for elements in $\vec x$ and $\vec y$. Afterwards, we reorder $\vec y$ by aligning each element to its corresponding same ranked element in $\vec x$. This actually reorganizes the style feature $\vec y$ according to the sorting order of the content feature $\vec x$. Then we reshape the adjusted feature to get $z \in \mathbb{R}^{C \cdot H \cdot W}$, which will be treated as the input of a decoder $D$ to generate the stylized image. The computing process of \abbr~is shown in Alg. \ref{alg}. 




We propose that the Style Projection module will not lose the color details of style images, while the structure of content images are also carried in the transferred features. In the following, we take insight into the style and content preservation achieved by Style Projection algorithm, respectively.

\noindent
\textbf{Style detail preservation.}
To understand the effectiveness of \methodname on style detail preservation, we investigate the Gram matrix of images. The Gram matrix $G(i,j)$ of feature map $y$ is formulated as
\begin{equation}
G(i,j)=\sum_{p}y(i,p)y(j,p)
\label{eq:gram}
\end{equation}
where $i$ and $j$ denote two channels on feature map $y$, and $p$ denotes the spatial position. Such that $G_y(i,j)$ is the inner product between two feature channels. 

The Gram matrix $G(i,j)$ can be used to evaluate texture synthesis algorithms by measuring the texture correlation between images \cite{gatys2016image}. Consider the difference between Gram matrices of feature maps $y$ and $z$,
\begin{align}
\begin{split}
\mathcal{L}_{Gram}(y,z) & = ||G_y-G_z||^2_{\rm{F}} \\
&=\sum_{i,j}||\sum_{p}y(i,p)y(j,p)-\sum_{p}z(i,p)z(j,p)||^2 
\end{split}
\label{eq:gramnotchange}
\end{align}
Gu et al. \cite{reshuffle} theoretically reveals that Eq. \ref{eq:gramnotchange} is equal to $0$ when $y$ and $z$ can be transformed to each other using a bijective transformation, for instance any shuffling function. Therefore, the feature shuffling does not alter the Gram matrix of feature. From Eq. \ref{eq:gramnotchange}, we show that the style information can be well preserved after style feature reshuffling in our proposed \methodname~module.

\begin{figure}[t]
\centering
\stackunder[3pt]{\includegraphics[width=0.24\linewidth]{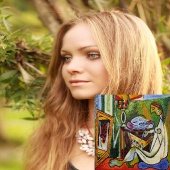}}{\footnotesize Content / Style}
\stackunder[3pt]{\includegraphics[width=0.24\linewidth]{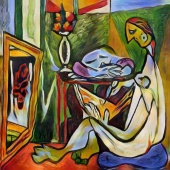}}{\footnotesize No Shuffling}
\stackunder[3pt]{\includegraphics[width=0.24\linewidth]{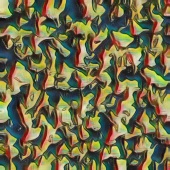}}{\footnotesize Random Shuffling}
\stackunder[3pt]{\includegraphics[width=0.24\linewidth]{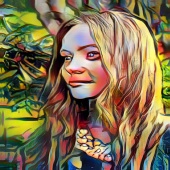}}{\footnotesize \abbr} \\
\vspace{-0.5em}
\caption{
An empirical study on feature shuffling mechanisms.
``No Shuffling'' directly feeds style features to the decoder without feeding content features or conducting any shuffling operation. 
``Random Shuffling'' feeds randomly shuffled style features (shuffled within each channel of feature maps) to the decoder. 
``Style Projection'' reorders style features according to order statistics of content features.
}
\label{fig:baseline}
\vspace{-0.5em}
\end{figure}

\noindent
\textbf{Content structure preservation.}
\methodname~also preserves the structure of content images. As suggested by \cite{david2004order}, the order of random variables contains effective information where the order statistics are related to the distribution function of random variables. In \methodname~module, the style features are reshuffled according to the order of content features, thus the structural relationships in content features can be implicitly inherited by the reshuffled style features.

To verify this claim, we conduct an empirical study on feature shuffling mechanisms and the results are shown in Fig. \ref{fig:baseline}. We observe that among the three methods, only \methodname~is capable of fusing content and style properly. `No shuffling' of style features shows an exact reconstruction of the original style image. `Random shuffling' shows a repetition of random style patterns, demonstrating that the style patterns can be preserved after feature reshuffling. Only \methodname, which reorders the style features based on content features, shows reasonable image stylization. This study reveals the effectiveness of \methodname~as it is able to preserve both content and style information during feature transformation. Through \methodname, the spatial structure (shape, edge, etc.) of content images and the color/texture patterns of style images are well fused.

\subsection{Style Transfer Model}

\begin{figure}[t]
    \centering
    \includegraphics[width=1\linewidth]{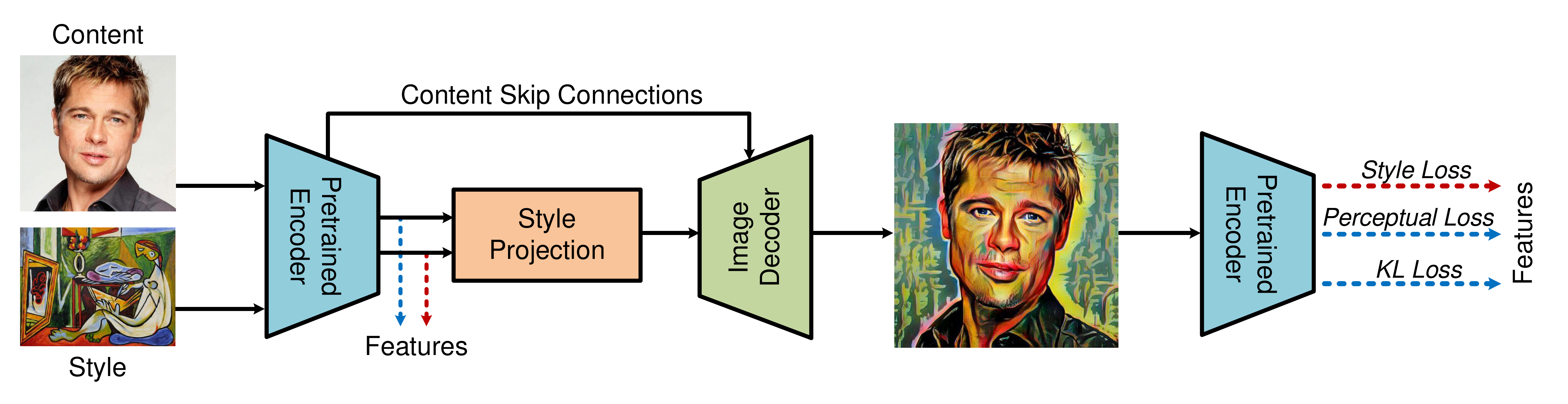}
    \vspace{-2em}
    \caption{The scheme of our proposed style transfer framework. 
    }
    \label{fig:framework}
    \vspace{-1em}
\end{figure}

We illustrate our style transfer model in Fig. \ref{fig:framework}. The learning objective of our style transfer model is composed of three parts, including the style loss, the content perceptual loss, and the content KL divergence loss. The style loss $\mathcal{L}_s$ is used to match the feature statistics between the style image $s$ and the stylized image $\hat{c}$ as $\mathcal{L}_s = \sum_{i=1}^{N} \lVert \mu,\sigma(E_{i}(s))-\mu,\sigma(E_{i}(\hat{c})) \rVert_{2}$,
where $\mu$ and $\sigma$ denotes the mean and standard deviation, and $E_i$ is the intermediate output of the $i$-th layer of encoder $E$, and $N$ is the number of encoder layers. The content perceptual loss $\mathcal{L}_p$ \cite{johnson2016perceptual} is used to minimize the pixel-wise feature distance between the content image $c$ and the stylized image $\hat{c}$ as $\mathcal{L}_p = \left \| E(c)-E(\hat{c})\right \|_{2}$.

Most image style transfer methods suffer from non-natural image clues in stylized results. We conjecture this is partially due to the missing of semantic information such as brightness. Inspired by the insights provided by generative models \cite{goodfellow2014generative}, we introduce a distribution matching objective, Kullback-Leibler (KL) divergence \cite{kl}, into the style transfer framework to leverage more semantic information in content images, as
\begin{equation}
    \mathcal{L}_{KL} = \mathcal{KL}\left[E(c)~||~E(\hat{c})\right]
\end{equation}
where $c$ is an input content image and $\hat{c}$ is the stylized image produced by decoder $D$. With the aid of KL divergence, we regularize $D$ to generate images that contain more semantics of content images. 

Overall, the complete learning objective of our model is formulated as $ \mathcal{L} = \mathcal{L}_p + \lambda \mathcal{L}_s + \kappa \mathcal{L}_{KL}$, where $\lambda$ and $\kappa$ are the loss weights of $\mathcal{L}_s$ and $\mathcal{L}_{KL}$, respectively. As \methodname~is a parameter-free method and the added KL divergence loss only brings negligible extra computing time, our style transfer approach is highly efficient.


\section{Experiments}

\subsection{Experimental Setup}


\noindent
\textbf{Datasets.}
We adopt the training set of MS-COCO dataset \cite{mscoco} as the content images and that of WikiArt dataset \cite{wikiart} as the style images. In training, we resize all input images to the size of 512$\times$512 and randomly crop each image to 256$\times$256. All content and style images used for testing purpose are selected from the test set of the two datasets, and the test images are never observed by the model during training. Our encoder and decoder networks work in a fully-convolutional manner, and they can be applied to images of arbitrary size in testing.

\noindent
\textbf{Implementation.}
We implement our style transfer model\footnote{\href{https://github.com/PaddlePaddle/PaddleHub/tree/dbca09ae78b5387ebe3b49f37ce88de45d41d26a/hub_module/modules/image/style_transfer/stylepro_artistic}{\textcolor{blue}{https://github.com/PaddlePaddle/PaddleHub/Stylepro\_Artistic}}} based on the PaddlePaddle framework \cite{ma2019paddlepaddle} and release a pretrained model as an official PaddleHub tool\footnote{\href{https://www.paddlepaddle.org.cn/hubdetail?name=stylepro\_artistic}{\textcolor{blue}{https://www.paddlepaddle.org.cn/hubdetail?name=stylepro\_artistic}}}.
We train our style transfer model for 160,000 iterations. We use an Adam optimizer \cite{kingma2014adam} with an initial learning rate of 1e-4 and a learning rate decay of 5e-5. Unless otherwise specified, the loss weights $\lambda$ and $\kappa$ are set to 10 and 2.5, respectively. Note that the loss weights are employed to balance the style transfer and content semantics preservation. More empirical studies on the KL divergence weight $\kappa$ are illustrated in the following ablation studies. We use a batch size of 8 for training.


\noindent
\textbf{Networks.}
The encoder network used in the model is the VGG-19 network \cite{vgg} pre-trained on ImageNet \cite{imagenet}, and its weights are not updated during training. The decoder network is composed of nine \textquoteleft Padding-Conv-ReLU\textquoteright~blocks, except the last block that has no ReLU layer. Three up-sampling layers are adopted right after the 1-st, 5-th, and 7-th block to restore the input image dimension successively, where the nearest neighbor interpolation is employed for up-sampling. We do not use normalization layer in our decoder network since it will hurt the diversity of synthesized images \cite{adain}. 

\begin{table}[!t]
\centering
\caption{Quantitative evaluation of style loss $\mathcal{L}_s$, content loss $\mathcal{L}_p$, and total loss. The lower the better.
}
\vspace{-1em}
\resizebox{1\hsize}{!}{
\begin{tabular}{l|c|c|c}
\hline
\textbf{Method} & \textbf{Style loss} $\mathcal{L}_s$ & \textbf{Content loss} $\mathcal{L}_p$ & \textbf{Total} \\
\hline
CNN \cite{gatys2016image} & 0.90  & 2.51 & 3.41 \\
StyleSwap \cite{styleswap} & 4.87 & 2.56 & 7.43 \\
WCT \cite{wct} & 1.33 & 4.08 & 5.41 \\
AdaIN \cite{adain} & 0.39 & 2.56 & 2.95 \\
LinearPropagation \cite{li2019learning} & 2.86 & 2.86 & 5.73\\
\hline
\abbr & \textbf{0.38} & 2.61 & 2.99 \\
\abbr~+ Skip & 0.48 & 2.56 & 3.04 \\
\abbr~+ KL & \textbf{0.38} & 2.25 & 2.63\\
\abbr~+ Skip + KL & 0.49 & \textbf{2.09} & \textbf{2.58} \\
\hline
\end{tabular}}

\label{table:loss}
\end{table}

\begin{figure*}[t]
\centering

\includegraphics[width=0.098\linewidth]{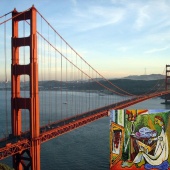}
\includegraphics[width=0.098\linewidth]{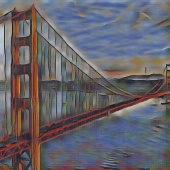}
\includegraphics[width=0.098\linewidth]{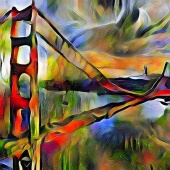}
\includegraphics[width=0.098\linewidth]{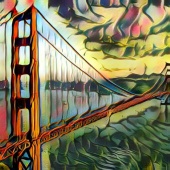}
\includegraphics[width=0.098\linewidth]{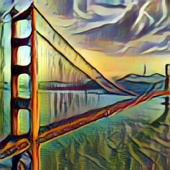}
\includegraphics[width=0.098\linewidth]{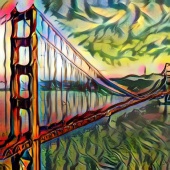}
\includegraphics[width=0.098\linewidth]{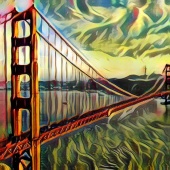}



\stackunder[5pt]{\includegraphics[width=0.098\linewidth]{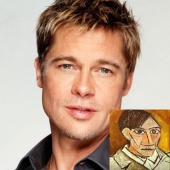}}{\small Content/Style}
\stackunder[5pt]{\includegraphics[width=0.098\linewidth]{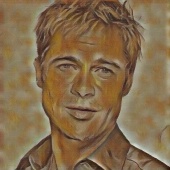}}{\small StyleSwap}
\stackunder[5pt]{\includegraphics[width=0.098\linewidth]{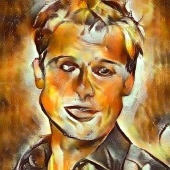}}{\small WCT}
\stackunder[5pt]{\includegraphics[width=0.098\linewidth]{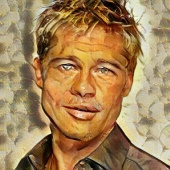}}{\small AdaIN}
\stackunder[5pt]{\includegraphics[width=0.098\linewidth]{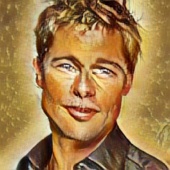}}{\small LinearProp}
\stackunder[5pt]{\includegraphics[width=0.098\linewidth]{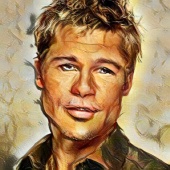}}{\small Ours-1}
\stackunder[5pt]{\includegraphics[width=0.098\linewidth]{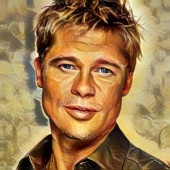}}{\small Ours-2}

\vspace{-0.5em}
\caption{%
Qualitative comparisons between the state-of-the-art content-style transformation modules. Ours-2 adds content skip connections and KL divergence loss to our Style Projection algorithm (Ours-1) to deliver more appealing images. 
%
}
\label{fig:comparison}
\vspace{-0.5em}
\end{figure*}

\subsection{Comparison with State-of-the-Arts}

\noindent
\textbf{Quantitative evaluation.} In Fig. \ref{fig:comparison}, we quantitatively evaluate the state-of-the-art arbitrary image style transfer methods, including Style-Swap \cite{styleswap}, WCT \cite{wct}, AdaIN \cite{adain}, Linear Propagation \cite{li2019learning}. We investigate the content loss $\mathcal{L}_p$, style loss $\mathcal{L}_s$ and total loss in inference. Our \abbr~shows competitive results compared with the state-of-the-art style transfer modules with respect to both style and content loss. The SP+Skip+KL method presents more superior results than the other methods and similar style loss as AdaIN, indicating that the modules proposed in this paper, including \abbr~and KL divergence loss, work well in arbitrary style transfer tasks. We also observe that there is a trade-off between style loss and content loss. Generally, SP+KL and SP+Skip+KL reach a good balance in such a trade-off.

\begin{figure}[t]
\centering
\stackunder[3pt]{\includegraphics[width=0.24\linewidth]{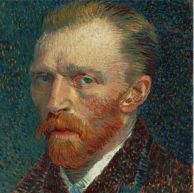}}{\small Content}
\hfill
\stackunder[3pt]{\includegraphics[width=0.24\linewidth]{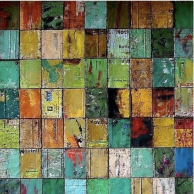}}{\small Style}  \hfill
\stackunder[3pt]{\includegraphics[width=0.24\linewidth]{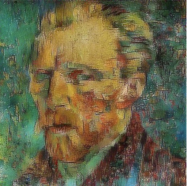}}{\small DFR \cite{reshuffle}}
\hfill
\stackunder[3pt]{\includegraphics[width=0.24\linewidth]{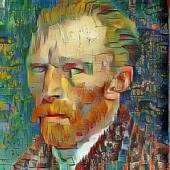}}{\small SP} \vspace{-0.5em}
\caption{
A comparison of shuffling-based transfer methods.
}

\label{fig:dfs}
\end{figure}

\noindent
\textbf{Qualitative comparison.} In Fig. \ref{fig:comparison} we show stylized images generated by different methods. The images generated by WCT \cite{wct} have a low fidelity probably due to the whitening operation which peels off some critical content clues. AdaIN \cite{adain} better reconstructs the content details, however there are still several \textquoteleft non-natural strokes\textquoteright~caused by only transferring the mean and standard deviation of the style features. The images produced by the StyleSwap \cite{styleswap} are too dark probably due to the way of replacing content features. Our methods (last two columns) provide more pleasant results. Even using \methodname~independently (\textit{i.e.}, Ours-1), style information is properly transferred while main content is preserved. SP+Skip+KL (\textit{i.e.}, Ours-2) produces more appealing results where content details and semantics are better preserved. 
Both DFR \cite{reshuffle} and SP are reshuffling-based feature transformation methods. A comparison between DFR and Style Projection is shown in Fig. \ref{fig:dfs}. DFR includes an expensive optimization process in its framework, thus it is time-consuming in image generation (taken about 114 seconds/image). Our parameter-free \methodname~is able to give appealing stylized results with a high efficiency (0.068 second/image). 


\begin{table}[t]
\caption{User study on style transfer methods. 
``\textbf{Method Score}'' is the percentage of times that users prefer the images in pairwise comparisons.
``SP'' denotes Style Projection and ``Full'' denotes Style Projection+Skip+KL.
``\textbf{User Consistency}'' denotes the consistency among users. 
``Scene'' and ``Person'' are two categories of content images.
}
\vspace{-1em}
\footnotesize
\centering
\begin{tabular}{l|c|c|c|c}
\hline
\multirow{2}{*}{}  & \multicolumn{3}{c|}{\textbf{Method Score} (\%)}                   & \multirow{2}{*}{\textbf{User Consistency} (\%)}  \\ \cline{2-4}
 & \multicolumn{1}{c|}{AdaIN} & \multicolumn{1}{c|}{SP} & Full  &   \\ \hline
Scene   & 32.58 & 51.89  & \textbf{65.53} & 68.94 \\ \hline
Person    & 34.47  & 37.50 & \textbf{78.03} & 76.52 \\ \hline 
\end{tabular}

\label{userstudy}
\end{table}





\subsection{User Study}
We further conduct a user study to evaluate three different methods, including AdaIN, our SP, and our full model (SP+Skip+KL). We build 24 pairs of synthesized images including 8 pairs of AdaIN and SP, 8 pairs of AdaIN and Full, and 8 pairs of SP and Full. The users are asked to choose one synthesized image from each pair under the criteria: (1) whether the image is sharp and clean while its content is close to that of the content image, and (2) whether the image has a similar style as the style image. We have collected feed-backs from 33 individuals and the results are summarized in Table \ref{userstudy}. The `Method Score' is the preference ratio averaged over all the image pairs and users, and the results indicate that most users favor the synthesized images of our methods. Interestingly, for the person images, more users prefer our full model (78.03\%), which indicates that fidelity is deserved to be well maintained for person images (see Fig. \ref{fig:comparison}).

\section{Conclusion}
In this paper, we have presented a real-time feed-forward model for arbitrary style transfer. The core is a parameter-free, fast, and effective feature-level style transformation algorithm named \methodname. We have also introduced the KL divergence loss into our style transfer model for a regularization of semantic consistency on content structures. Extensive experiments including quantitative evaluation, qualitative analysis, and user study have validated the efficacy of our method for arbitrary image style transfer.


\small
\bibliographystyle{IEEEbib}
\bibliography{main}

\end{document}


\newcommand{\capname}{Style Projection}
\newcommand{\name}{Style Projection}
\newcommand{\abbr}{Style Projection}

\acmConference[ACM MM '20]{ACM MM '20: 28th ACM International Conference on Multimedia}{Oct 12--16, 2020}{Seattle, United States}
\acmBooktitle{ACM MM '20: 28th ACM International Conference on Multimedia,
  Oct 12--16, 2020, Seattle, United States}
\acmPrice{}
\acmISBN{}

\title{Parameter-Free Style Projection for Arbitrary Style Transfer
\\Supplementary Material}
\author{ACM MM '20 Submission ID 119}

\twocolumn[{%
\renewcommand\twocolumn[1][]{#1}%
\maketitle
\begin{center}

\includegraphics[width=0.135\linewidth]{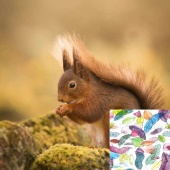}
\includegraphics[width=0.135\linewidth]{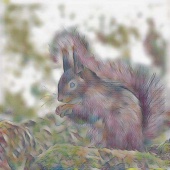}
\includegraphics[width=0.135\linewidth]{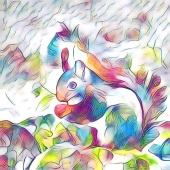}
\includegraphics[width=0.135\linewidth]{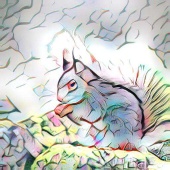}
\includegraphics[width=0.135\linewidth]{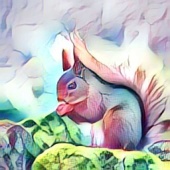}
\includegraphics[width=0.135\linewidth]{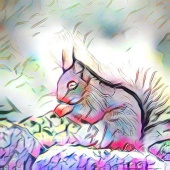}
\includegraphics[width=0.135\linewidth]{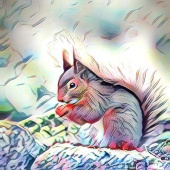}
\vspace{0.1cm}

\includegraphics[width=0.135\linewidth]{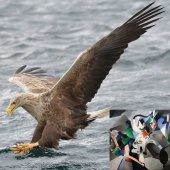}
\includegraphics[width=0.135\linewidth]{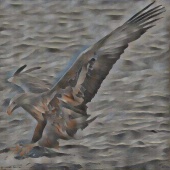}
\includegraphics[width=0.135\linewidth]{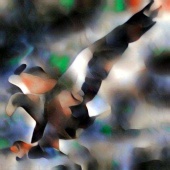}
\includegraphics[width=0.135\linewidth]{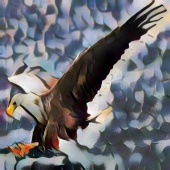}
\includegraphics[width=0.135\linewidth]{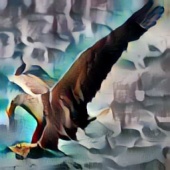}
\includegraphics[width=0.135\linewidth]{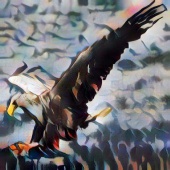}
\includegraphics[width=0.135\linewidth]{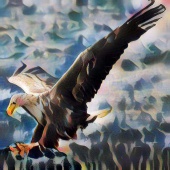}
\vspace{0.1cm}

\includegraphics[width=0.135\linewidth]{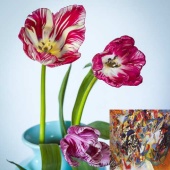}
\includegraphics[width=0.135\linewidth]{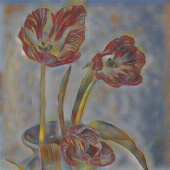}
\includegraphics[width=0.135\linewidth]{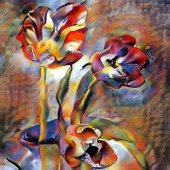}
\includegraphics[width=0.135\linewidth]{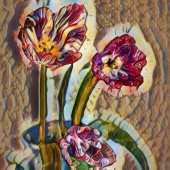}
\includegraphics[width=0.135\linewidth]{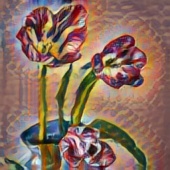}
\includegraphics[width=0.135\linewidth]{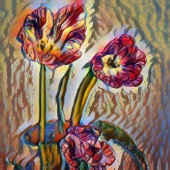}
\includegraphics[width=0.135\linewidth]{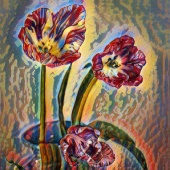}
\vspace{0.1cm}

\includegraphics[width=0.135\linewidth]{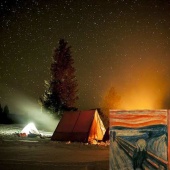}
\includegraphics[width=0.135\linewidth]{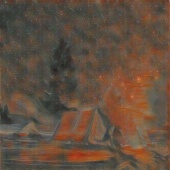}
\includegraphics[width=0.135\linewidth]{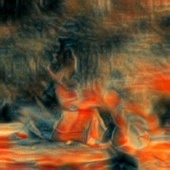}
\includegraphics[width=0.135\linewidth]{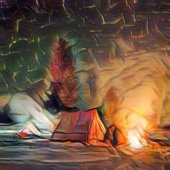}
\includegraphics[width=0.135\linewidth]{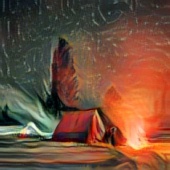}
\includegraphics[width=0.135\linewidth]{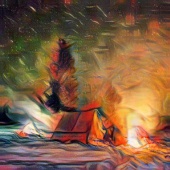}
\includegraphics[width=0.135\linewidth]{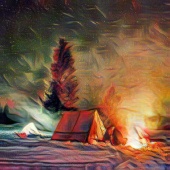}
\vspace{0.1cm}

\stackunder[5pt]{\includegraphics[width=0.135\linewidth]{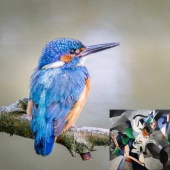}}{\small Content / Style}
\stackunder[5pt]{\includegraphics[width=0.135\linewidth]{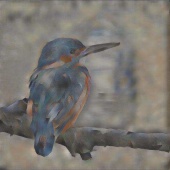}}{\small Style Swap \cite{styleswap}}
\stackunder[5pt]{\includegraphics[width=0.135\linewidth]{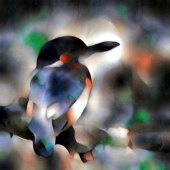}}{\small WCT \cite{wct}}
\stackunder[5pt]{\includegraphics[width=0.135\linewidth]{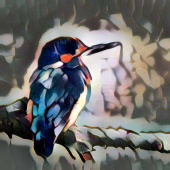}}{\small AdaIN \cite{adain}}
\stackunder[5pt]{\includegraphics[width=0.135\linewidth]{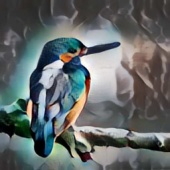}}{\small Linear Prop \cite{li2019learning}}
\stackunder[5pt]{\includegraphics[width=0.135\linewidth]{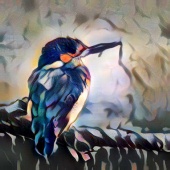}}{\small Ours}
\stackunder[5pt]{\includegraphics[width=0.135\linewidth]{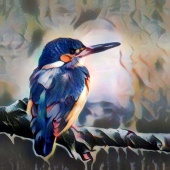}}{\small Ours+Skip+KL}

\captionof{figure}{A comparison between the state-of-the-art feed-forward style transfer methods, as a supplement to Fig. 7 of the paper.}
\end{center}
}]

\begin{figure*}[t]
\centering

\includegraphics[width=0.135\linewidth]{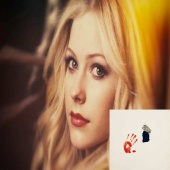}
\includegraphics[width=0.135\linewidth]{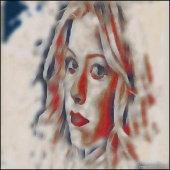}
\includegraphics[width=0.135\linewidth]{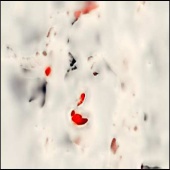}
\includegraphics[width=0.135\linewidth]{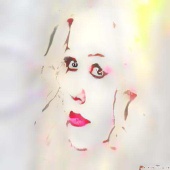}
\includegraphics[width=0.135\linewidth]{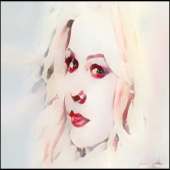}
\includegraphics[width=0.135\linewidth]{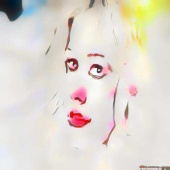}
\includegraphics[width=0.135\linewidth]{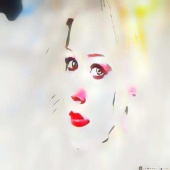}
\vspace{0.1cm}

\includegraphics[width=0.135\linewidth]{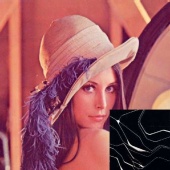}
\includegraphics[width=0.135\linewidth]{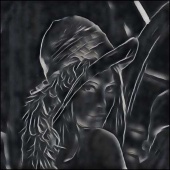}
\includegraphics[width=0.135\linewidth]{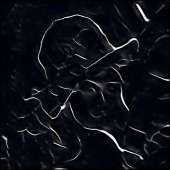}
\includegraphics[width=0.135\linewidth]{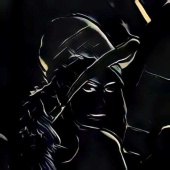}
\includegraphics[width=0.135\linewidth]{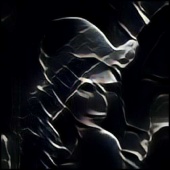}
\includegraphics[width=0.135\linewidth]{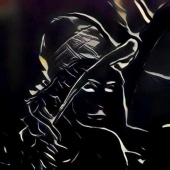}
\includegraphics[width=0.135\linewidth]{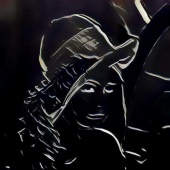}
\vspace{0.1cm}

\includegraphics[width=0.135\linewidth]{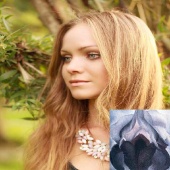}
\includegraphics[width=0.135\linewidth]{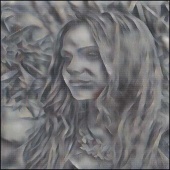}
\includegraphics[width=0.135\linewidth]{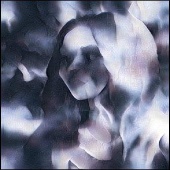}
\includegraphics[width=0.135\linewidth]{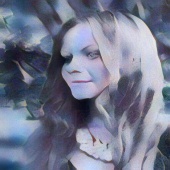}
\includegraphics[width=0.135\linewidth]{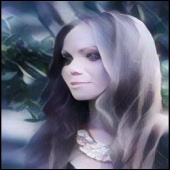}
\includegraphics[width=0.135\linewidth]{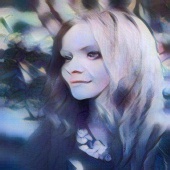}
\includegraphics[width=0.135\linewidth]{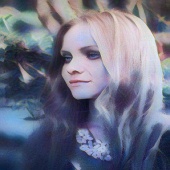}
\vspace{0.1cm}

\includegraphics[width=0.135\linewidth]{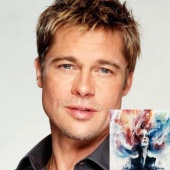}
\includegraphics[width=0.135\linewidth]{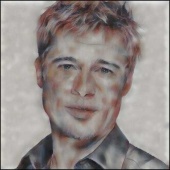}
\includegraphics[width=0.135\linewidth]{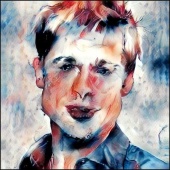}
\includegraphics[width=0.135\linewidth]{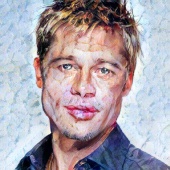}
\includegraphics[width=0.135\linewidth]{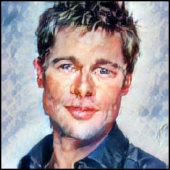}
\includegraphics[width=0.135\linewidth]{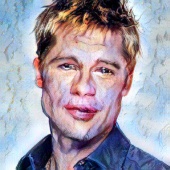}
\includegraphics[width=0.135\linewidth]{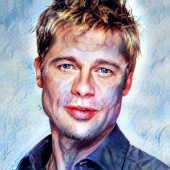}
\vspace{0.1cm}

\includegraphics[width=0.135\linewidth]{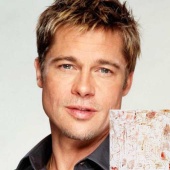}
\includegraphics[width=0.135\linewidth]{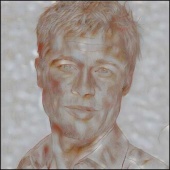}
\includegraphics[width=0.135\linewidth]{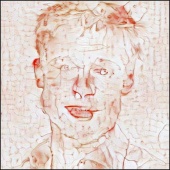}
\includegraphics[width=0.135\linewidth]{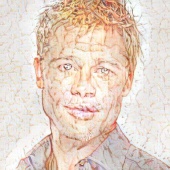}
\includegraphics[width=0.135\linewidth]{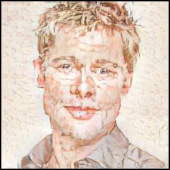}
\includegraphics[width=0.135\linewidth]{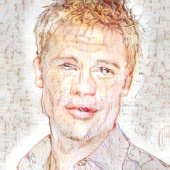}
\includegraphics[width=0.135\linewidth]{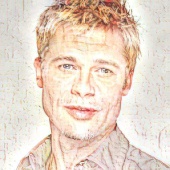}
\vspace{0.1cm}

\includegraphics[width=0.135\linewidth]{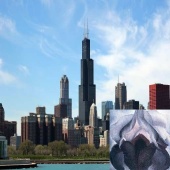}
\includegraphics[width=0.135\linewidth]{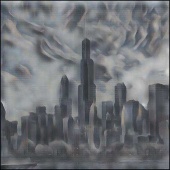}
\includegraphics[width=0.135\linewidth]{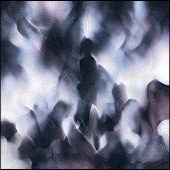}
\includegraphics[width=0.135\linewidth]{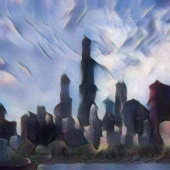}
\includegraphics[width=0.135\linewidth]{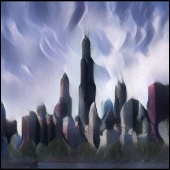}
\includegraphics[width=0.135\linewidth]{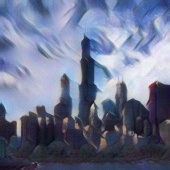}
\includegraphics[width=0.135\linewidth]{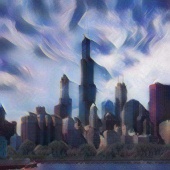}
\vspace{0.1cm}

\stackunder[5pt]{\includegraphics[width=0.135\linewidth]{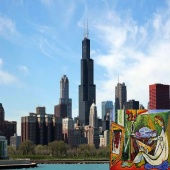}}{\small Content / Style}
\stackunder[5pt]{\includegraphics[width=0.135\linewidth]{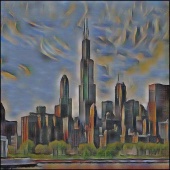}}{\small Style Swap \cite{styleswap}}
\stackunder[5pt]{\includegraphics[width=0.135\linewidth]{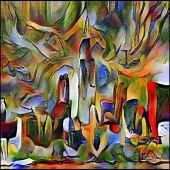}}{\small WCT \cite{wct}}
\stackunder[5pt]{\includegraphics[width=0.135\linewidth]{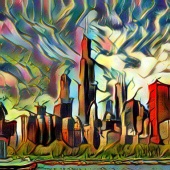}}{\small AdaIN \cite{adain}}
\stackunder[5pt]{\includegraphics[width=0.135\linewidth]{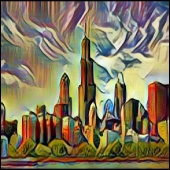}}{\small Linear Prop \cite{li2019learning}}
\stackunder[5pt]{\includegraphics[width=0.135\linewidth]{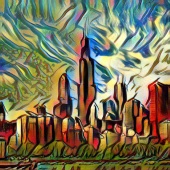}}{\small Ours}
\stackunder[5pt]{\includegraphics[width=0.135\linewidth]{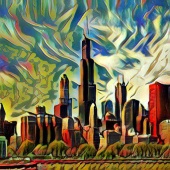}}{\small Ours+Skip+KL}
\caption{%
A comparison between the state-of-the-art feed-forward style transfer methods, as a supplement to Fig. 7 of the paper.
}
\label{fig:comparison1}
\end{figure*}

\begin{figure*}[t]
\centering

\includegraphics[width=0.135\linewidth]{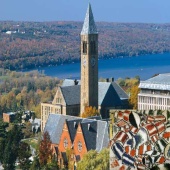}
\includegraphics[width=0.135\linewidth]{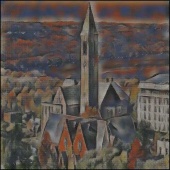}
\includegraphics[width=0.135\linewidth]{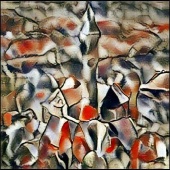}
\includegraphics[width=0.135\linewidth]{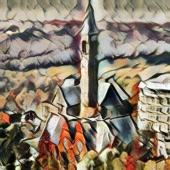}
\includegraphics[width=0.135\linewidth]{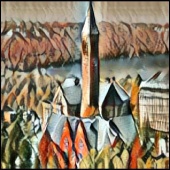}
\includegraphics[width=0.135\linewidth]{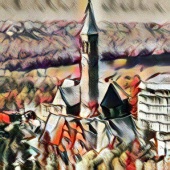}
\includegraphics[width=0.135\linewidth]{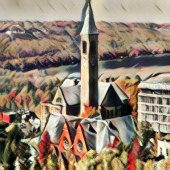}
\vspace{0.1cm}

\includegraphics[width=0.135\linewidth]{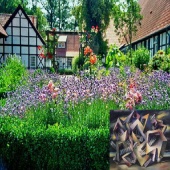}
\includegraphics[width=0.135\linewidth]{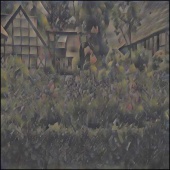}
\includegraphics[width=0.135\linewidth]{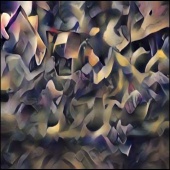}
\includegraphics[width=0.135\linewidth]{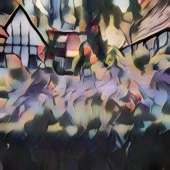}
\includegraphics[width=0.135\linewidth]{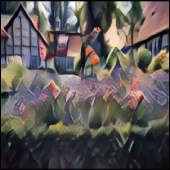}
\includegraphics[width=0.135\linewidth]{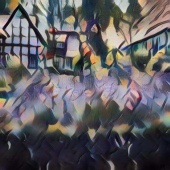}
\includegraphics[width=0.135\linewidth]{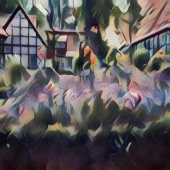}
\vspace{0.1cm}

\includegraphics[width=0.135\linewidth]{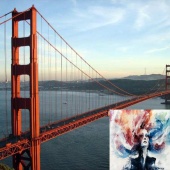}
\includegraphics[width=0.135\linewidth]{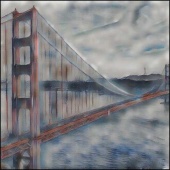}
\includegraphics[width=0.135\linewidth]{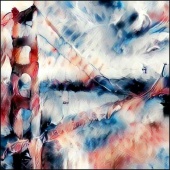}
\includegraphics[width=0.135\linewidth]{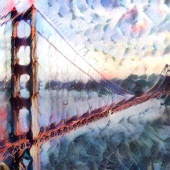}
\includegraphics[width=0.135\linewidth]{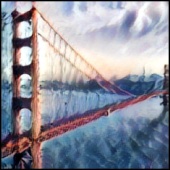}
\includegraphics[width=0.135\linewidth]{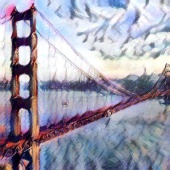}
\includegraphics[width=0.135\linewidth]{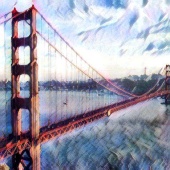}
\vspace{0.1cm}

\includegraphics[width=0.135\linewidth]{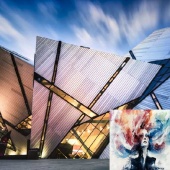}
\includegraphics[width=0.135\linewidth]{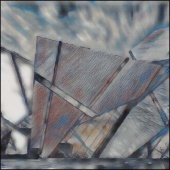}
\includegraphics[width=0.135\linewidth]{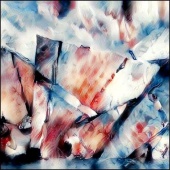}
\includegraphics[width=0.135\linewidth]{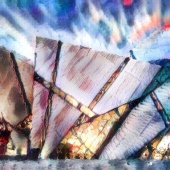}
\includegraphics[width=0.135\linewidth]{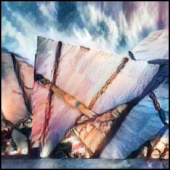}
\includegraphics[width=0.135\linewidth]{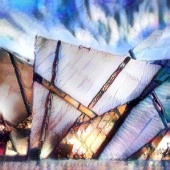}
\includegraphics[width=0.135\linewidth]{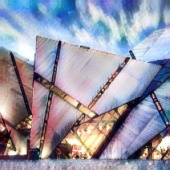}
\vspace{0.1cm}

\includegraphics[width=0.135\linewidth]{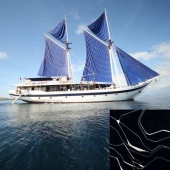}
\includegraphics[width=0.135\linewidth]{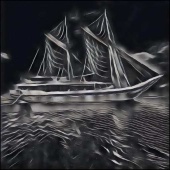}
\includegraphics[width=0.135\linewidth]{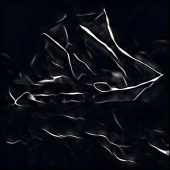}
\includegraphics[width=0.135\linewidth]{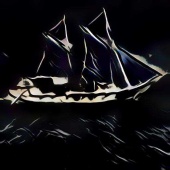}
\includegraphics[width=0.135\linewidth]{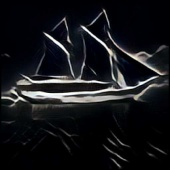}
\includegraphics[width=0.135\linewidth]{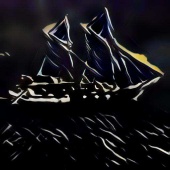}
\includegraphics[width=0.135\linewidth]{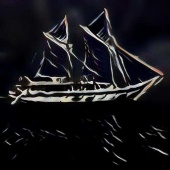}
\vspace{0.1cm}

\includegraphics[width=0.135\linewidth]{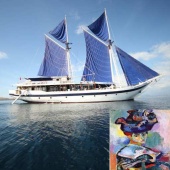}
\includegraphics[width=0.135\linewidth]{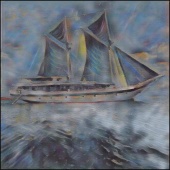}
\includegraphics[width=0.135\linewidth]{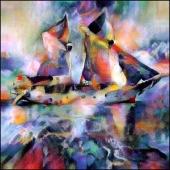}
\includegraphics[width=0.135\linewidth]{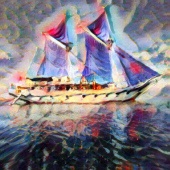}
\includegraphics[width=0.135\linewidth]{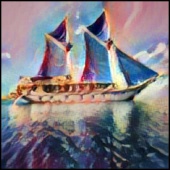}
\includegraphics[width=0.135\linewidth]{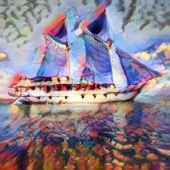}
\includegraphics[width=0.135\linewidth]{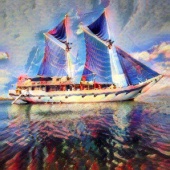}
\vspace{0.1cm}

\stackunder[5pt]{\includegraphics[width=0.135\linewidth]{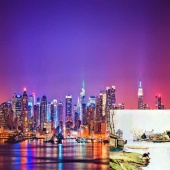}}{\small Content / Style}
\stackunder[5pt]{\includegraphics[width=0.135\linewidth]{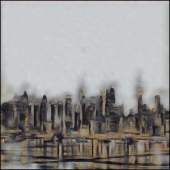}}{\small Style Swap \cite{styleswap}}
\stackunder[5pt]{\includegraphics[width=0.135\linewidth]{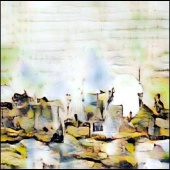}}{\small WCT \cite{wct}}
\stackunder[5pt]{\includegraphics[width=0.135\linewidth]{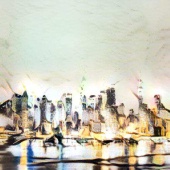}}{\small AdaIN \cite{adain}}
\stackunder[5pt]{\includegraphics[width=0.135\linewidth]{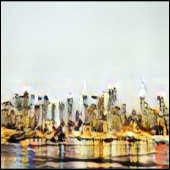}}{\small Linear Prop \cite{li2019learning}}
\stackunder[5pt]{\includegraphics[width=0.135\linewidth]{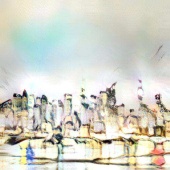}}{\small Ours}
\stackunder[5pt]{\includegraphics[width=0.135\linewidth]{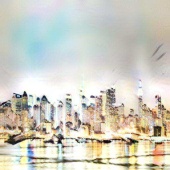}}{\small Ours+Skip+KL}
\caption{%
A comparison between the state-of-the-art feed-forward style transfer methods, as a supplement to Fig. 7 of the paper.
}
\label{fig:comparison1}
\end{figure*}

\clearpage
{
\bibliographystyle{ACM-Reference-Format}
\bibliography{ref}
}